\documentclass[11pt]{article}

\usepackage[preprint]{acl}

\usepackage{times}
\usepackage{latexsym}
\usepackage{multirow}

\usepackage[T1]{fontenc}

\usepackage[utf8]{inputenc}

\usepackage{microtype}

\usepackage{inconsolata}

\usepackage{graphicx}
\usepackage{subcaption}
\usepackage{amsmath}
\usepackage{booktabs}
\title{ReCache: Efficient KV Cache Reuse and Compression for Tool-Augmented LLM Agents}

\author{
 \textbf{Yichu Fang\textsuperscript{1,2}},
 \textbf{Sitong Wei\textsuperscript{3}},
 \textbf{Haozhe Hu\textsuperscript{1,2}},
 \textbf{Xiaoyu Shen\textsuperscript{2}\thanks{Corresponding author}}
\\
 \textsuperscript{1}Shanghai Jiao Tong University\\
 \textsuperscript{2}Eastern Institute of Technology, Ningbo\\
 \textsuperscript{3}Xi'an Jiaotong University
\\
\href{mailto:fy_chu@sjtu.edu.cn}{fy\_chu@sjtu.edu.cn}
 \href{mailto:xyshen@eitech.edu.cn}{xyshen@eitech.edu.cn}
}

\begin{document}
\maketitle
\begin{abstract}
Agentic language models repeatedly encode tool and skill schemas that recur across requests in different combinations and orders, preventing standard prefix caching from reusing their key--value (KV) states. We introduce \textbf{ReCache}, a framework for independently caching resource representations while reducing their inference-time computational and memory overhead. Resource-wise attention removes cross-resource interactions and assigns resource-local positions, producing composition-invariant KV blocks. ReCache then restricts resource visibility to contribution-selected layer--KV-head-group routes and retains only invocation-critical fields through structural and semantic pruning. We evaluate ReCache on a benchmark assembled from seven public tool- and skill-use datasets, including resource-disjoint tests. Resource-wise attention matches dense invocation performance (82.3\% versus 82.4\% Inv-F1) while providing a 3.655$\times$ time-to-first-token speedup. The complete framework reduces allocated KV-tensor memory by 92.43\% and accelerates attention by 1.423$\times$. These results show that separating reusable schema encoding from selective resource access substantially reduces agentic inference costs with limited effectiveness loss. The code is available at \url{https://github.com/EIT-NLP/ReCache}.
\end{abstract}

\section{Introduction}
\begin{figure}[ht]
\centering
\begin{subfigure}[t]{0.45\textwidth}
    \includegraphics[width=\textwidth]{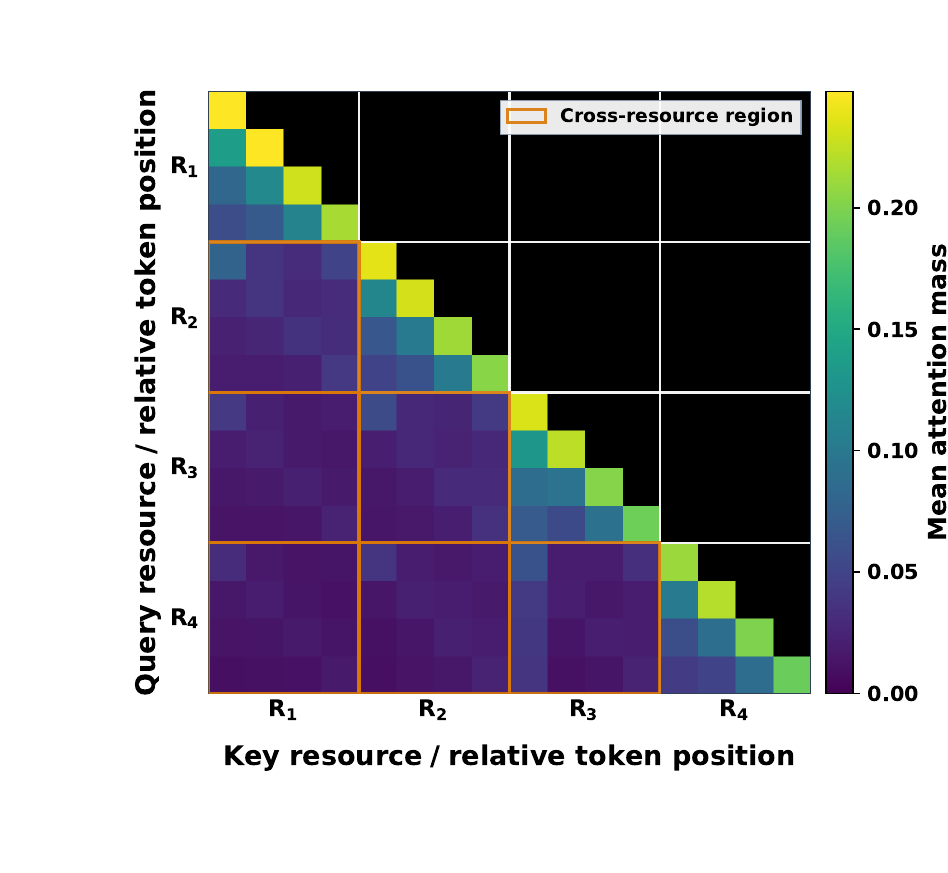}
    \caption{\small Causal attention among resources (relative-position bins). Orange boxes mark cross-resource blocks.}
    \label{fig:intro-map-4b}
\end{subfigure}
\hfill
\begin{subfigure}[t]{0.48\textwidth}
    \includegraphics[width=\textwidth]{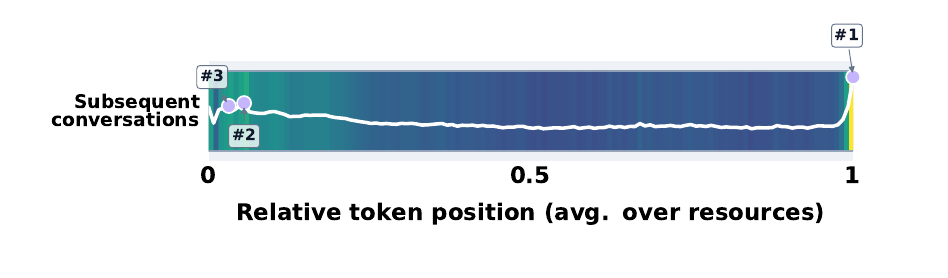}
    \caption{\small Attention from subsequent conversation tokens onto resources, normalized and aligned by relative position.}
    \label{fig:intro-context-4b}
\end{subfigure}
\caption{\small Cross-resource attention on Qwen3-4B.}
\label{fig:intro}
\end{figure}


Recent advances have extended large language models (LLMs) into the core reasoning module of agentic systems, where they address user requests by dynamically invoking external tools and skills~\citep{schick2023toolformer,liu2024toolace,li2026skillsbench,xu2026agent}. As the set expands, agentic systems commonly adopt \emph{progressive disclosure} and retrieve only task-relevant schemas during execution~\citep{toolret2025,zheng2026skillrouter}. We refer to each tool or skill schema as a \emph{resource}. Although retrieval limits the active context, the same resource may recur across requests in different combinations and orders, causing repeated prefill computation for unchanged information.

KV-cache reuse provides a natural way to amortize this repeated computation. However, standard prefix caching requires reused content to form an identical prefix, which rarely holds for dynamically composed resource contexts~\citep{ye2025kvcomm,zhang2026sparsex}. Recent approaches extend KV reuse to modular or position-independent contexts~\citep{gim2024prompt,hu2024epic,yao2025cacheblend,yang2025kvlink}, but often require positional correction or selective recomputation to approximate full-context KV states. Meanwhile, transformers can jointly encode semantic and positional dependencies~\citep{vaswani2017attention,shaw2018self,su2024roformer}, and the importance of positional information varies across tasks, with some studies showing that certain tasks can maintain performance under reduced or altered positional signals~\citep{wang2021on,haviv2022transformer,kazemnejad2023impact}. In resource invocation, preserving resource-internal semantics may therefore matter more than preserving global resource order. Supporting this intuition, Figure~\ref{fig:intro-map-4b} shows that Qwen3-4B exhibits substantially stronger attention within individual resources than across resources.

Based on this observation, we introduce \textbf{ReCache}, a resource-level KV reuse and compression framework for dynamically retrieved tools and skills. Its core mechanism, \emph{resource-wise attention}, removes attention between distinct resource blocks and resets positional indices within each block. The resulting KV representation is independent of neighboring resources and their absolute positions, enabling resource caches to be constructed and reused independently. Although this representation differs from conventional full-context prefill, lightweight fine-tuning adapts the model to the resulting attention and positional patterns while preserving reliable resource invocation.

While resource-wise attention enables independent KV reuse, storing and transmitting complete resource caches can still be costly as resource length and scale grow. ReCache therefore compresses resource caches from both \emph{structural} and \emph{semantic} perspectives. Structurally, transformer layers and KV head groups contribute unevenly to model predictions~\citep{michel2019sixteen,voita2019analyzing,dehghanighobadi2026depthkv}. ReCache ranks layers and head groups by their marginal contribution to prediction loss and retains resource visibility only along the most important routes. This contribution-based criterion exploits the distinct sparsity patterns across sequential layers and parallel KV head groups, providing a more reliable measure of resource utility than attention-mass selection. Semantically, Figure~\ref{fig:intro-context-4b} shows that the final suffix token receives the highest attention mass from subsequent conversation tokens, suggesting that it naturally aggregates information from preceding fields~\citep{muennighoff2022sgpt}. This observation motivates retaining the suffix token as a compact semantic anchor, similar in spirit to summary-based compression methods that aggregate prompt or segment information into summary tokens~\citep{mu2023learning,chevalier2023adapting,zhang2024long}. However, unlike general text, tool and skill schemas must preserve exact interface information, including resource identifiers, argument names, and parameter constraints; invalid names and arguments are common sources of invocation failure~\citep{song2023restgpt,qin2024toolllm,hao2023toolkengpt}. ReCache therefore retains these critical fields alongside the suffix token for semantic aggregation.

To evaluate resource reuse and generalization systematically, we construct a unified tool- and skill-use benchmark from seven public datasets. We first apply a diversity-first sampling strategy to maximize resource diversity across samples, followed by filtering of invalid traces containing hallucinated resource usages. The resulting benchmark contains both in-distribution and resource-disjoint out-of-distribution splits, enabling evaluation on both known and unseen resources.

Experiments with Qwen3 backbones show that resource-wise attention differs from Dense, the standard dense-attention baseline, by at most 0.2\% across effectiveness metrics while providing a 3.655$\times$ time-to-first-token (TTFT) speedup. With structural and semantic pruning, ReCache reduces allocated KV-tensor memory by 92.43\% and accelerates attention by 1.423$\times$. Under matched structural budgets, contribution-based selection preserves higher invocation effectiveness than attention-based and layer-asymmetric alternatives, particularly on unseen resources. Direct comparisons further show that field-aware semantic pruning retains better performance than generic text-compression baselines under the evaluated configurations. Across increasing resource lengths, ReCache maintains nearly constant TTFT, TPOT, and attention latency while capping allocated KV-tensor memory at 0.03 GiB.

Our contributions are threefold: (1) \textbf{ReCache} provides a resource-level KV reuse and compression framework for dynamically retrieved tools and skills; (2) resource-wise attention, contribution-guided structural pruning, and field-aware semantic pruning jointly reduce redundancy in cache construction, layer--head-group access, and token retention; and (3) a unified tool- and skill-use benchmark with resource-disjoint out-of-distribution evaluation supports systematic analysis of efficiency and invocation performance.

\section{Related Work}
\paragraph{Agentic Resources.}
Tool-augmented language models invoke external functions described by structured schemas. Recent benchmarks also evaluate function calling over large API pools and multi-turn interactions \citep{li2023apibank,liu2024toolace,qin2024toolllm,liu2024apigen,hammerbench2024}. Agent systems also use natural-language skills that package reusable procedures with invocation metadata \citep{xu2026agent,sok2026agenticskills,li2026skillsbench}. As these resource pools grow, the corresponding retrieval selects a task-relevant subset before inference \citep{toolret2025,zheng2026skillrouter}. Retrieval reduces the number of resources in the prompt, but it does not eliminate repeated encoding of recurring resources.

\paragraph{Reusable Caching.}

Prefix caching reuses KV states for requests sharing an identical prefix~\citep{kwon2023efficient,zheng2024sglang,yu2025stateful}. 
However, its applicability to dynamically composed resources remains limited, as reordered or recombined segments may differ from the positional configurations under which their KV states were constructed. Position-independent methods address this mismatch through various strategies. EPIC~\citep{hu2024epic} and CacheBlend~\citep{yao2025cacheblend} preserve original positional encodings while selectively recomputing salient tokens to improve cache integration. In particular, CacheBlend further corrects positional information across cached KV states to mitigate position mismatch during reuse. KVLink~\citep{yang2025kvlink} instead decouples positional information from cached content and compensates for positional shifts during reuse. KVCOMM~\citep{ye2025kvcomm}, on the other hand, uses the entropy of embedding-based anchor weights to assess cache shareability. Nevertheless, the necessity of explicit positional information can vary across tasks. NoPo~\citep{haviv2022transformer} and NoPE~\citep{kazemnejad2023impact} demonstrate that Transformers can recover positional knowledge without explicit positional encodings in various downstream tasks, while Wang et al.~\citep{wang2021on} further show that positional information exhibits task-dependent importance across classification and span prediction.


\paragraph{KV-Cache Compression.}

KV-cache compression exploits structural redundancy across tokens, layers, and attention heads. H2O evicts low-importance states under a uniform layer budget, whereas DepthKV assigns layer-dependent retention budgets~\citep{zhang2023h2o,dehghanighobadi2026depthkv}. DuoAttention classifies heads by their attention roles, and SPEED restricts prompt-KV visibility by depth while preserving full-depth decoding states~\citep{xiao2025duoattention,oh2026shallow}. These studies establish the importance of multidimensional sparsity, but their selection criteria are not optimized for resource invocation. In addition, semantic compression reduces the number of retained context tokens. Learned approaches encode spans into compact summary states~\citep{mu2023learning,chevalier2023adapting,zhang2024long}, while LLMLingua and LongLLMLingua apply query-aware token filtering~\citep{jiang2023llmlingua,jiang2024longllmlingua}. ChunkKV further retains contiguous segments to preserve local coherence~\citep{liu2026chunkkv}. These techniques effectively compress natural-language contexts while retaining salient information, but resources contain structured identifiers and field relationships that directly determine invocation behavior, motivating compression strategies tailored to field-level semantics.

\begin{figure*}[htbp]
    \centering
    \includegraphics[width=\textwidth]{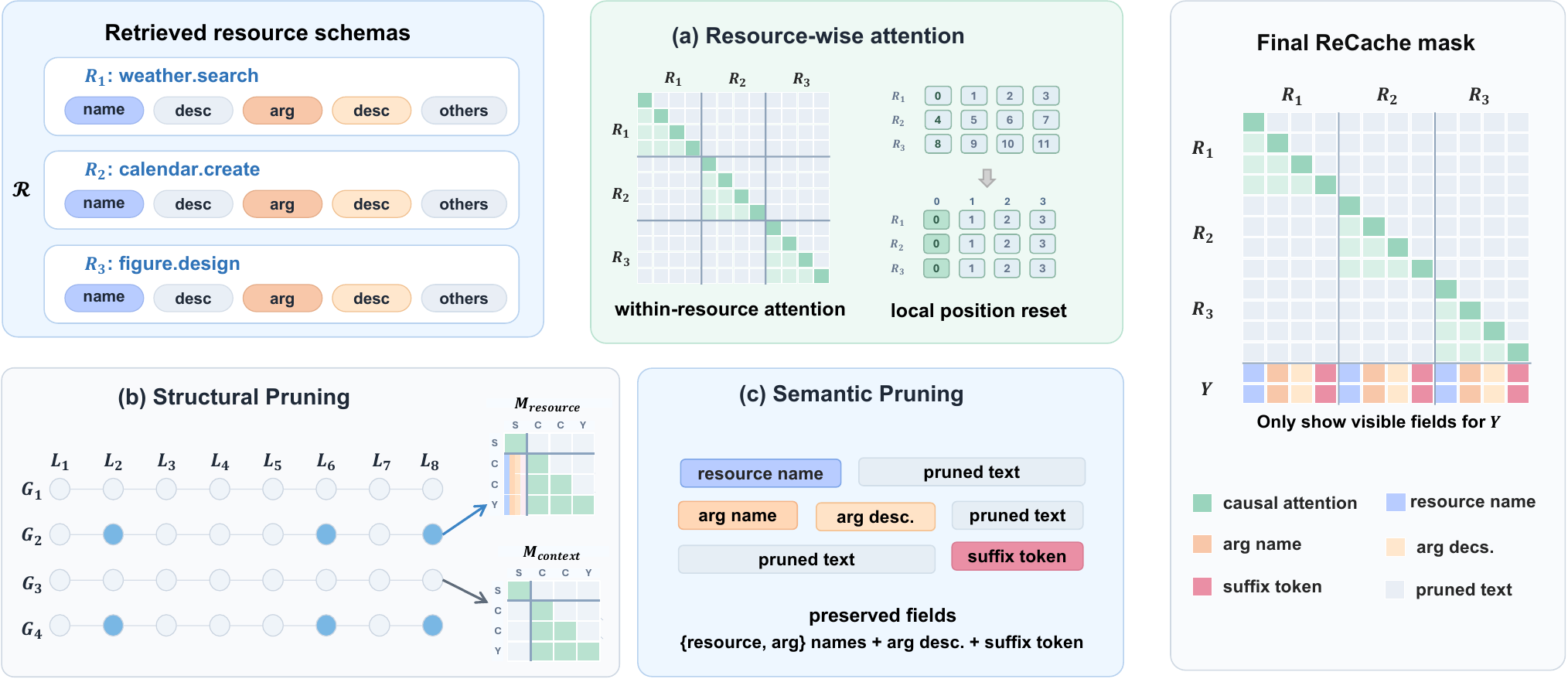}
    \caption{\small ReCache constructs reusable KV blocks for each $R_i$ and progressively reduces their subsequent cache retention and access. (a) Resource-wise attention removes inter-resource attention and resets positions from zero within each $R_i$. (b) Structural pruning exposes resource KV states (using $M_{\mathrm{resource}}$) only through selected layer ($L_i$)--KV-head-group ($G_i$) routes $\Omega^\star$ (highlighted in blue). The remaining routes use $M_{\mathrm{context}}$. (c) Semantic pruning retains resource names, argument (arg) names and descriptions (desc.), and the final suffix token. The final mask combines these decisions, ensuring that $Y$ attends exclusively to retained semantic fields within assigned routes during decoding.}
    \label{fig:03arch}
\end{figure*}
\section{ReCache}
\paragraph{Problem Setup}
\label{sec:method-setup}
Given a dataset $\mathcal{D}$, each instance $(X,Y)\in\mathcal{D}$ consists of an agentic input sequence $X$ and the target resource invocations $Y$ required to satisfy a user request, including tool and skill invocations. The input sequence comprises the system instruction, user query, retrieved resources $\mathcal{R}=\{R_1,R_2,\ldots,R_N\}$, and preceding conversational history when available in multi-turn interactions. Each resource $R_i=(t_{i,1},\ldots,t_{i,D_i})$ is represented as a token sequence of length $D_i$ describing its functionality, including resource identifiers, argument definitions, and associated metadata.

The objective is to predict the required invocations $Y$ conditioned on $X$. As illustrated in Figure~\ref{fig:03arch}, ReCache improves resource-processing efficiency through three progressive stages. Resource-wise attention removes unnecessary resource-resource interactions to enable independent KV representations; structural pruning restricts resource visibility to selected layer--KV-head-group routes; and semantic pruning reduces the visible resource length by retaining only critical fields.

\paragraph{Resource-Wise Attention}
\label{sec:method-rwa}
Retrieval-Augmented Generation (RAG) demonstrates that disjoint retrieved passages can be encoded independently without mutual cross-attention~\cite{izacard2021leveraging,lin2025refrag}. Extending this principle, agentic resources are also retrieved \cite{toolret2025} and primarily rely on internal semantics, allowing their KV caches to be constructed and reused independently of retrieval combinations or positional orderings.

Defining $D_{\mathcal{R}}=\sum_iD_i$ as the total resource length, resource-wise attention removes cross-resource dependencies during cache construction. As illustrated in Figure~\ref{fig:03arch}(a), tokens within each resource attend only to the shared prefix and their local resource tokens, while attention links between different resources are eliminated. The resource--resource attention is thus reduced from $O(D_{\mathcal{R}}^2)$ to $O(\sum_iD_i^2)$, allowing each $R_i$ to be independently represented and cached.

To achieve full reusability, resource KV representations must also remain positionally invariant. Following prior studies on task-dependent positional importance~\citep{wang2021on,haviv2022transformer,kazemnejad2023impact}, we hypothesize that absolute placement among retrieved resources provides limited benefit for resource invocation. Therefore, we adopt a resource-local positional layout, where each resource follows an identical relative position scheme that preserves internal token continuity while removing dependence on global resource ordering. Specifically, during cache construction, ReCache assigns resource-local positional indices $pos(t_{i,j})=j$ while preserving the original positional encodings of the surrounding context. Consequently, each resource receives identical positional embeddings regardless of its placement among retrieved items. While frameworks like EPIC \citep{hu2024epic} incorporate position resetting as part of a broader compilation pipeline, empirical ablations confirm that this re-indexing is sufficient to preserve task effectiveness.


\paragraph{Structural Pruning}
\label{sec:method-stc}
\label{sec:method-structural}

Transformer layers and attention heads exhibit substantial redundancy~\citep{michel2019sixteen,voita2019analyzing,men2024shortgpt}. ReCache exploits this structural sparsity by restricting resource--context visibility to influential layer--KV-head-group routes. Following objective-driven selection methods~\citep{xiao2025duoattention,dehghanighobadi2026depthkv}, route importance is measured by its marginal reduction in invocation loss.

Formally, let $[L]=\{1,\ldots,L\}$ and $[G]=\{1,\ldots,G\}$ denote the sets of transformer layers and KV head groups under grouped-query attention (GQA), respectively. A structural routing configuration is defined by a subset $\Omega\subseteq[L]\times[G]$, where each $(l,g)\in\Omega$ represents a layer--head-group route with access to resource KV states constructed by resource-wise attention. As illustrated in Figure~\ref{fig:03arch}(b), routes in $\Omega$ use $M_{\mathrm{resource}}$ to attend to the retained resource tokens and the conversational context. Routes outside $\Omega$ use $M_{\mathrm{context}}$, which structurally prunes all resource tokens from their attention scope while preserving standard causal attention over non-resource tokens.

The key challenge is to identify a compact $\Omega$ without degrading task performance. Contrary to DepthKV \citep{dehghanighobadi2026depthkv}, which evaluates layer importance by pruning one layer at a time from a dense configuration, ReCache applies leave-one-in analysis to both layers and KV head groups\footnote{In preliminary experiments, leave-one-out scoring produced only small loss differences between structural units, whereas attention-mass ranking yielded lower invocation effectiveness than contribution-based selection. The latter result may stem from heterogeneous sensitivity across structural dimensions and token positions~\citep{ding2024longrope}.}. Starting from $\Omega=\emptyset$, where conversation tokens cannot access resource states, it measures the reduction in invocation loss obtained by activating one layer or head group.

We denote the held-out selection set by $\mathcal{D}_{\mathrm{h}}$ and the span of the target response $Y$ by $\mathcal{T}(Y)$. For each $(X,Y)\in\mathcal{D}_{\mathrm{h}}$, the invocation loss under configuration $\Omega$ is

\begin{equation*}
\ell_{\Omega}(X,Y)=-\frac{1}{|\mathcal{T}(Y)|}\sum_{t\in\mathcal{T}(Y)} \log p_{\theta}^{\Omega}(y_t\mid y_{<t},X),
\end{equation*}

where $p_{\theta}^{\Omega}$ denotes the model distribution induced by $\Omega$. Restricting the objective to invocation-related positions aligns route selection with resource identification and argument generation, while preventing unrelated response tokens from dominating the score. The selection objective $\mathcal{J}(\Omega)$ is then the average of $\ell_{\Omega}(X,Y)$ over $\mathcal{D}_{\mathrm{h}}$.

For layer $l$, we activate all associated KV head groups using $\Omega_l=\{l\}\times[G]$ and define the layer contribution as

\begin{equation*}
s_l=\mathcal{J}(\emptyset)-\mathcal{J}(\Omega_l).
\end{equation*}

A positive $s_l$ indicates that resource visibility at layer $l$ improves prediction performance relative to $\Omega=\emptyset$. Head-group contributions $s_g$ are computed analogously using $\Omega_g=[L]\times\{g\}$. For each structural dimension, positive contributions are normalized into contribution weights:

\begin{equation*}
w_i=\frac{\max(s_i,0)}{\sum_j\max(s_j,0)}
\end{equation*}

where $i$ denotes a layer or KV head group. A larger $w_i$ indicates a greater isolated contribution to resource-dependent prediction, and guides the selection of candidate structural budgets.

Specifically, given layer and head-group budgets $K_L$ and $K_G$, ReCache retains the highest-weight units according to $w_i$:
\begin{align*}
\mathcal{L}^{\star} &=\operatorname{TopK}_{K_L}(\{w_l\}), \\
\mathcal{G}^{\star} &=\operatorname{TopK}_{K_G}(\{w_g\}),
\end{align*}

and constructs the final routing configuration as

\begin{equation*}
\Omega^\star=\mathcal{L}^{\star}\times\mathcal{G}^{\star}.
\end{equation*}

During inference, only routes in $\Omega^\star$ expose resource KV states to conversational texts to reduce unnecessary computation.

\paragraph{Semantic Pruning}
\label{sec:method-smc}
While resource-wise attention enables resource-level cache reuse and structural pruning reduces unnecessary resource visibility, each activated route still processes the full $D_i$ tokens within independent KV blocks. These tokens are inherently structural \cite{qin2024toolllm,bai2026skillzip} and span heterogeneous fields, including executable identifiers, parameter constraints, descriptions, and formatting elements, with varying contributions to invocation performance, motivating field-aware compression. Since resource invocation failures are predominantly associated with invalid names and incorrect arguments~\cite{song2023restgpt, qin2024toolllm, hao2023toolkengpt}, ReCache performs semantic pruning along the token dimension to preserve invocation-critical components while removing redundant content.

For each $R_i$, ReCache retains tokens from three fields: the resource name, argument names, and argument descriptions. These fields support resource identification, parameter-interface alignment, and argument-value constraints, respectively. Ablation results show that preserving these three fields incurs only minor performance degradation\footnote{Preliminary experiments with attention-based top-$k$ token selection (while preserving resource and argument names) provide limited improvements, suggesting that complete argument descriptions contain essential semantic information. Chunk-based selection strategies following~\citet{zhang2024long,chevalier2023adapting} require substantially more retained segments while providing limited interpretability.}.

Additionally, the final token is preserved as a suffix representation, whose hidden state aggregates preceding resource information. This design is related to summary-based compression approaches~\citep{mu2023learning,zhang2024long}, but instead of introducing new summary tokens, it reuses the final token of $R_i$ as a naturally occurring boundary representation whose causal state can condition on all preceding fields~\citep{muennighoff2022sgpt,wang2024improving,zhuang2024promptreps}\footnote{Preserving multiple trailing tokens (e.g., the last 10 or 20 tokens) provides limited additional gains.}. 
The resulting subsequence $\widetilde{R}_i$ has an effective length of $\widetilde{D}_i\leq D_i$.

The resulting compact KV representations remain independently reusable, while conversational texts on routes in $\Omega^\star$ attend only to the retained tokens. As illustrated by the \textit{Final ReCache Mask} in Figure~\ref{fig:03arch}, this selective visibility further reduces resource--context attention overhead.

\section{Experimental Setup}
\subsection{Datasets}

We evaluate ReCache on a unified resource-use dataset covering both tools and skills. Since existing datasets typically focus on particular resource types or interaction patterns, we combine seven sources to broaden resource
and scenario coverage, including ToolACE~\citep{liu2024toolace}, APIGEN~\citep{liu2024apigen}, ToolMind~\citep{yang2025toolmind}, ToolRet~\citep{toolret2025}, Toucan~\citep{toucan2025}, SkillRouter~\citep{zheng2026skillrouter}, and WildToolBench~\citep{yu2026benchmarking}.

Direct aggregation, however, introduces several confounding factors. Synthetic traces may contain malformed invocations or use resources absent from the provided set. More subtly, resource knowledge can be encoded into model parameters through learned embeddings or special tokens \citep{hao2023toolkengpt,wang2025toolgen}. ToolOmni further observes that parameter memorization generalizes poorly to unseen and evolving resources \citep{huang2026toolomni}. Frequently repeated resource names or exact candidate subsets may consequently allow the model to recall familiar interfaces rather than interpret the schema supplied at inference time. We therefore retain only parseable and grounded resource calls and adopt \textbf{diversity-first sampling}, which prioritizes newly observed called resources and candidate-set configurations before satisfying source-level quotas.\footnote{An initial direct-aggregation strategy exhibited strong frequency bias toward recurring resources and duplicated candidate subsets. Random downsampling failed to alleviate this imbalance, motivating the two-stage diversity-first selection detailed in Appendix~\ref{sec:app-dataset}.}

The final training dataset $\mathcal{D}$ contains 49,424 examples, while a subset $\mathcal{D}_{\mathrm{subset}}$ with 5,000 examples is used for preliminary experiments. Evaluation is conducted on two test sets with 1,000 examples each: an in-distribution split $\mathcal{T}_{\mathrm{IND}}$, where the involved resources are observed during training, and an out-of-distribution split $\mathcal{T}_{\mathrm{OOD}}$, where the resources are unseen in $\mathcal{D}$.

\subsection{Evaluation Metrics}
\paragraph{Effectiveness} We evaluate effectiveness using the turn-averaged resource-invocation F1 (\textbf{Inv-F1}), where each turn may involve multiple parallel resource invocations for one instance. This metric measures whether the model can generate the complete set of required invocations for a given query. A predicted invocation is considered correct only if both the resource name and all associated arguments exactly match the corresponding gold invocation. Beyond end-to-end performance, we further analyze model behavior from two complementary perspectives. Resource identification (Resource ID) is evaluated using precision (\textbf{ID-P}), recall (\textbf{ID-R}), and turn-averaged F1 (\textbf{ID-F1}) over resource names, reflecting the ability to identify relevant resources. We additionally report the resource hallucination rate (\textbf{Halluc.}), defined as the proportion of predicted resource names that do not exist in the available resource set, which captures failures caused by generating invalid or unavailable resources. All metrics are reported in percentages (\%).

\paragraph{Efficiency.} We use time-to-first-token (\textbf{TTFT}) to quantify online prefill latency under a cache hit, including request-local KV materialization but excluding offline cache construction. We further report time-per-output-token (\textbf{TPOT}) to measure end-to-end decoding latency, and attention latency (\textbf{Attn.}) to expose the decoding gains attributable specifically to reduced attention computation, which can be obscured in TPOT by unchanged projection and feed-forward costs. Finally, \textbf{Mem.} measures the resource reduction in memory allocated to KV-cache tensors.

\subsection{Backbone Models}

We perform full fine-tuning using Qwen3 variants \cite{yang2025qwen3}, with Qwen3-4B ($Q_l$) as the primary backbone. Qwen3-1.7B ($Q_s$) serves as an auxiliary backbone for analyzing the sensitivity of structural budgets to model scale. We train the primary Qwen3-4B model on an NVIDIA A800 GPU and run the preliminary Qwen3-1.7B experiments on an NVIDIA RTX 5000 GPU.

\section{Structural Budget Analysis}
Rather than treating structural budgets as unexplained hyperparameters, we derive them from how resource information is used across the network. Motivated by the non-uniform structural sensitivity observed in layer and KV-cache pruning \citep{men2024shortgpt,dehghanighobadi2026depthkv}, we use the leave-one-in scores from Section~\ref{sec:method-structural} to profile layers and KV head groups, then validate the resulting budgets through end-to-end ablations, compared with the full-visibility configuration $\Omega_{\mathrm{full}}$ after applying resource-wise attention. For $\Omega=\mathcal{L}\times\mathcal{G}$, structural sparsity is

\begin{equation*}
\rho(\Omega)=1-\frac{|\mathcal{L}||\mathcal{G}|}{LG},
\end{equation*}

which measures the fraction of routes without resource visibility. We also introduce contribution coverage (CC) to quantify the preserved structural capacity under a routing budget. CC is computed as the cumulative contribution weight $w_i$ of the retained top-$k$ layers or KV head groups.

\begin{figure}[t]
  \centering
  \includegraphics[width=0.4\textwidth]{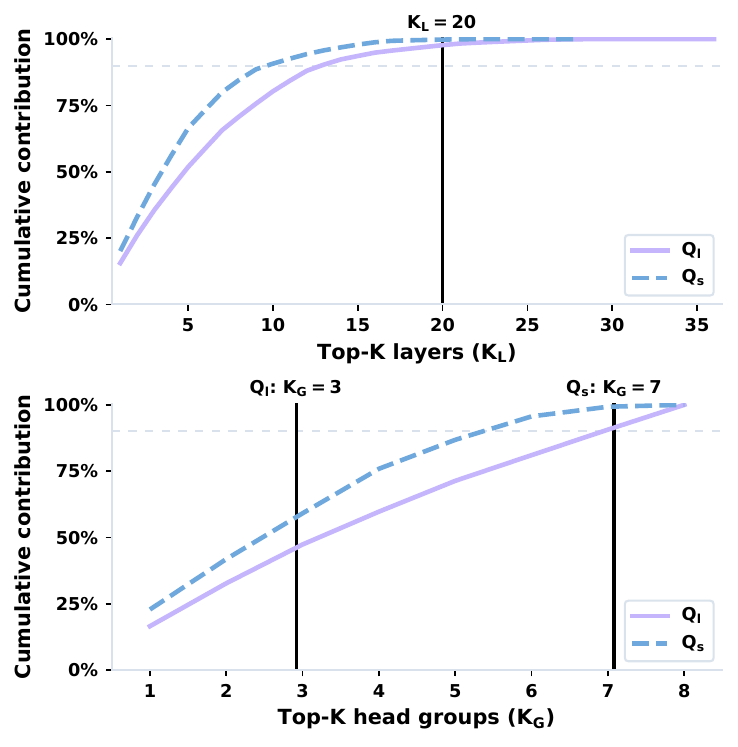}
  \caption{Structural contribution coverage in $Q_l$ and $Q_s$. Cumulative leave-one-in contributions of layers (top) and KV head groups (bottom); vertical lines mark the selected budgets.}
  \label{fig:structural-budget}
\end{figure}

\paragraph{Layer contributions.}
Figure~\ref{fig:structural-budget} shows that layer contributions are highly
concentrated: both cumulative curves rise rapidly and saturate near $K_L=20$, where the retained layers capture 97.7\% and 99.8\% CC for $Q_l$ and $Q_s$, respectively. This non-uniformity agrees with prior layer-pruning analyses \citep{fan2020layerdrop,men2024shortgpt} and indicates that exposing resource states in every layer is unnecessary. We therefore use $K_L=20$ as the shared layer budget and jointly validate it with the selected head groups
against $\Omega_{\mathrm{full}}$.

\paragraph{Head-group contributions.}
Head-group profiles differ by model scale. In $Q_l$, contributions are relatively flat, yet three groups retain near-full invocation performance despite covering only 47.3\% CC. This discrepancy suggests that resource-related functions are not concentrated in a single head group but distributed across multiple groups, allowing certain groups to be removed with limited impact~\citep{michel2019sixteen,voita2019analyzing}. In contrast, $Q_s$ requires seven groups, which cover 99.3\% CC and recover Inv-F1 to within 0.7 points of $\Omega_{\mathrm{full}}$. The smaller model therefore distributes this computation across more parallel groups, whereas the larger model offers greater substitutability, echoing the functional specialization of KV heads observed in long-context retrieval
\citep{wu2024retrievalhead,xiao2025duoattention}. Table~\ref{tab:structural-budget} summarizes the resulting budgets, structural sparsity, and invocation performance relative to $\Omega_{\mathrm{full}}$.

\begin{table}[ht]
\centering
\small
\setlength{\tabcolsep}{3.2pt}
\resizebox{\columnwidth}{!}{%
\begin{tabular}{lrrrrrr}
\toprule
Model & $K_L/L$ & $K_G/G$ & CC$_L$ & CC$_G$ & $\rho$ & $\Delta$Inv-F1 (\%) \\
\midrule
$Q_l$ & 20/36 & 3/8 & 97.7 & 47.3 & 79.2 & $-0.2$ \\
$Q_s$ & 20/28 & 7/8 & 99.8 & 99.3 & 37.5 & $-0.7$ \\
\bottomrule
\end{tabular}
}
\caption{Selected structural budgets. CC and $\rho$ are percentages; $\Delta$Inv-F1 is measured relative to $\Omega_{\mathrm{full}}$ after
full-data training.}
\label{tab:structural-budget}
\end{table}

\paragraph{Budget choice.}
We select the sparsest configuration that maintains performance comparable to $\Omega_{\mathrm{full}}$, obtaining $\Omega^\star=\Omega_{20,3}$ for $Q_l$ and $\Omega^\star=\Omega_{20,7}$ for $Q_s$.  Both configurations preserve nearly all cumulative layer contributions while assigning different head-group budgets according to model-specific contribution distributions\footnote{Appendix~\ref{sec:app-budget} presents detailed contribution weights and a preliminary budget analysis across model scales for identifying the optimal routing configuration $\Omega^\star$.}.

\section{Results}
All experiments in this section use $Q_l$ trained on the full training set $\mathcal{D}$ and are evaluated on $\mathcal{T}_{\mathrm{IND}}$ unless otherwise specified.

\paragraph{Cache Reuse}
Table~\ref{tab:rwa-effectiveness} reports both the effectiveness and efficiency of resource-wise attention. We compare against Dense, and evaluate two incremental configurations: Block eliminates cross-resource attention while retaining the original positional indices, and $\Omega_{\mathrm{full}}$ further reindexes positions to construct composition-independent resource blocks. Both configurations remain comparable to Dense, with all effectiveness metrics differing by at most 0.2\%. Resource-wise attention also enables resource representations to be pre-encoded with a shared prefix and reused across requests, reducing online computation and allowing $\Omega_{\mathrm{full}}$ to achieve a 3.655$\times$ TTFT speedup over Dense. Consistent with findings from other tasks~\citep{wang2021on,haviv2022transformer,kazemnejad2023impact}, these results suggest that cross-resource interactions and global resource ordering provide limited benefits in this setting, while resource-local position re-indexing preserves prediction quality and enables efficient resource-level reuse.

\begin{table}[ht]
\centering
\small
\setlength{\tabcolsep}{2pt}
\renewcommand{\arraystretch}{0.95}
\begin{tabular}{lcccc}
\toprule
\textbf{Method} & \textbf{Inv-F1} & \textbf{ID-F1} & \textbf{Halluc.} & \textbf{TTFT (ms)} \\
\midrule
Dense & 82.4 & 96.0 & 0.0 & 26.319 \\
Block & 82.2 & 95.8 & 0.1 & -- \\
$\Omega_{\mathrm{full}}$ & 82.3 & 96.0 & 0.2 & 7.200 \\
\bottomrule
\end{tabular}
\caption{\small Resource-wise attention ablation.}
\label{tab:rwa-effectiveness}
\end{table}

\paragraph{Model Comparison}
\begin{table}
    \centering
    \scriptsize
    \setlength{\tabcolsep}{4pt}
    \renewcommand{\arraystretch}{1.0}
    \begin{tabular}{@{}lccccc@{}}
    \toprule
    \textbf{Method}
    & \textbf{Attn. $\boldsymbol{\times}$} & \textbf{Mem. $\downarrow$}
    & \textbf{Inv-F1} & \textbf{ID-F1} & \textbf{Halluc.} \\
    \midrule
    Dense
    & -- & -- & 82.4 & 96.0 & 0.0 \\
    $\Omega_{\mathrm{full}}$
    & 1.001$\times$ & 0.47\% & 82.3 & 96.0 & 0.2 \\
    \midrule
    \multicolumn{6}{c}{\textit{Structural pruning}} \\
    \midrule
    $\Omega_{20,G}$
    & 1.016$\times$ & 44.71\% & 82.5 & 95.7 & 0.3 \\
    $\Omega_{\mathrm{full}}$ + SPEED
    & 1.013$\times$ & 44.71\% & 79.3 & 92.9 & 2.8 \\
    $\Omega_{20,3}$
    & 1.314$\times$ & 79.27\% & 82.1 & 95.6 & 0.2 \\
    $\Omega_{\mathrm{full}}$ + $\mathrm{SA}_{20,3}$
    & 1.302$\times$ & 79.27\% & 79.1 & 93.7 & 1.0 \\
    \midrule
    \multicolumn{6}{c}{\textit{Semantic pruning}} \\
    \midrule
    $\Omega_{\mathrm{full}}$ + Gist
    & 1.029$\times$ & 99.22\% & 39.2 & 46.9 & 51.4 \\
    $\Omega_{\mathrm{full}}$ + Beacon
    & 1.020$\times$ & 75.42\% & 78.6 & 92.5 & 1.4 \\
    $\Omega_{\mathrm{full}}$ + SMP
    & 1.021$\times$ & 63.57\% & 81.6 & 95.2 & 0.2 \\
    \midrule
    \textbf{ReCache}
    & \textbf{1.423$\times$} & 92.43\%
    & 80.3 & 94.9 & 0.2 \\
    \bottomrule
    \end{tabular}
    \caption{\small
    Results on $\mathcal{T}_{\mathrm{IND}}$.
    Efficiency is reported relative to Dense. SMP denotes Semantic Pruning.}
    \label{tab:main-ind}
\end{table}

\begin{table}
    \centering
    \scriptsize
    \setlength{\tabcolsep}{3pt}
    \renewcommand{\arraystretch}{1.0}
    \begin{tabular}{@{}lccc@{}}
    \toprule
    \textbf{Method}
    & \textbf{Inv-F1} & \textbf{ID-F1} & \textbf{Halluc.} \\
    \midrule
    Dense
    & 66.3 & 95.6 & 0.0 \\
    $\Omega_{\mathrm{full}}$
    & 64.7 & 94.4 & 0.4 \\
    \midrule
    \multicolumn{4}{c}{\textit{Structural pruning}} \\
    \midrule
    $\Omega_{20,G}$
    & 64.3 & 94.3 & 0.5 \\
    $\Omega_{\mathrm{full}}$ + SPEED
    & 54.3 & 81.6 & 13.4 \\
    $\Omega_{20,3}$
    & 63.2 & 93.3 & 0.5 \\
    $\Omega_{\mathrm{full}}$ + $\mathrm{SA}_{20,3}$
    & 58.2 & 88.4 & 5.1 \\
    \midrule
    \multicolumn{4}{c}{\textit{Semantic pruning}} \\
    \midrule
    $\Omega_{\mathrm{full}}$ + Gist
    & 9.7 & 17.8 & 79.9 \\
    $\Omega_{\mathrm{full}}$ + Beacon
    & 58.5 & 89.7 & 4.2 \\
    $\Omega_{\mathrm{full}}$ + SMP
    & 62.8 & 94.1 & 0.3 \\
    \midrule
    \textbf{ReCache}
    & 60.8 & 92.8 & 0.6 \\
    \bottomrule
    \end{tabular}
    \caption{\small $\mathcal{T}_{\mathrm{OOD}}$ Results. SMP denotes Semantic Pruning.}
    \label{tab:main-ood}
\end{table}

Table~\ref{tab:main-ind} compares structural and semantic pruning strategies built on $\Omega_{\mathrm{full}}$ on $\mathcal{T}_{\mathrm{IND}}$, reporting both efficiency and invocation effectiveness. To assess robustness to distribution shift, Table~\ref{tab:main-ood} reports effectiveness on $\mathcal{T}_{\mathrm{OOD}}$.

Under matched structural budgets, contribution-based selection consistently surpasses the corresponding baselines. At the 20-layer budget, $\Omega_{20,G}$ and $\Omega_{\mathrm{full}}+\mathrm{SPEED}$~\citep{oh2026shallow} exhibit nearly identical efficiency, whereas $\Omega_{20,G}$ improves Inv-F1 by 3.2\% on $\mathcal{T}_{\mathrm{IND}}$ and 10.0\% on $\mathcal{T}_{\mathrm{OOD}}$. For joint layer--head-group pruning, $\mathrm{SA}_{20,3}$ ranks routes by their attention mass over each resource and retains the top 20 layers and 3 KV head groups. Despite comparable efficiency, $\Omega_{20,3}$ improves Inv-F1 over $\mathrm{SA}_{20,3}$ by 3.0\% and 5.0\% points on the two test sets, respectively. The OOD hallucination rates further decrease from 13.4\% to 0.5\% relative to SPEED and from 5.1\% to 0.5\% relative to $\mathrm{SA}_{20,3}$, demonstrating the stronger generalization of contribution-based routing.

For semantic pruning\footnote{For fair comparison, both baselines are adapted to the resource setting by assigning summary tokens to each resource, following their original token-level compression mechanism.}, Gist~\citep{mu2023learning} attains the largest standalone KV reduction but severely degrades invocation effectiveness, indicating that a single summary vector cannot preserve the fine-grained information required for resource invocation. Beacon~\citep{zhang2024long} retains more information through chunk-level summaries, yet remains less effective than field-aware Semantic Pruning and exhibits higher hallucination, particularly on $\mathcal{T}_{\mathrm{OOD}}$. This gap indicates that partial coverage of critical fields is insufficient for robust generalization to unseen resources. Detailed field ablations are provided in Appendix~\ref{sec:app-sm}.

Combining all strategies, \textbf{ReCache} achieves the highest attention speedup (1.423$\times$) and a comparable KV reduction (92.43\%) among configurations that retain competitive effectiveness. It retains Inv-F1 scores of 80.3\% on $\mathcal{T}_{\mathrm{IND}}$ and 60.8\% on $\mathcal{T}_{\mathrm{OOD}}$, corresponding to 97.5\% and 91.8\% of performance of Dense, respectively.

\paragraph{Efficiency Scaling}
\begin{figure*}[ht]
    \centering
    \begin{subfigure}[t]{0.4\textwidth}
        \includegraphics[width=\textwidth]{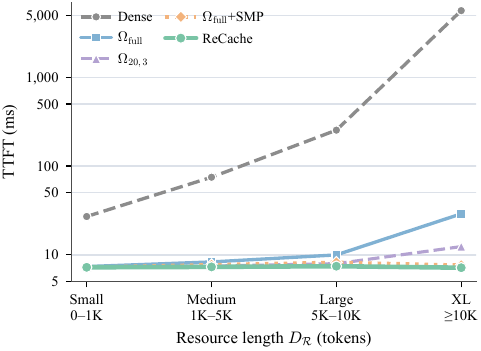}
        \caption{\small TTFT}
        \label{fig:res-eff-ttft}
    \end{subfigure}
    \hfill
    \begin{subfigure}[t]{0.4\textwidth}
        \includegraphics[width=\textwidth]{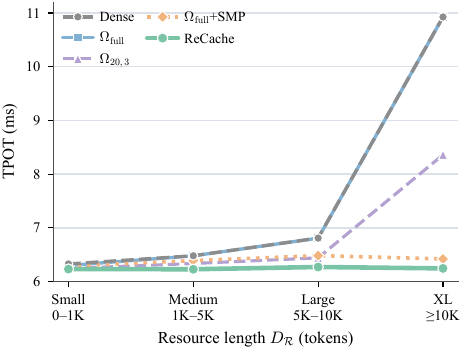}
        \caption{\small TPOT}
        \label{fig:res-eff-tpot}
    \end{subfigure}
    \hfill
    \begin{subfigure}[t]{0.4\textwidth}
        \includegraphics[width=\textwidth]{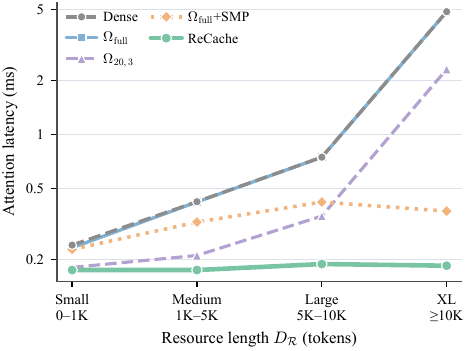}
        \caption{\small Attn.}
        \label{fig:res-eff-attn}
    \end{subfigure}
    \hfill
    \begin{subfigure}[t]{0.4\textwidth}
        \includegraphics[width=\textwidth]{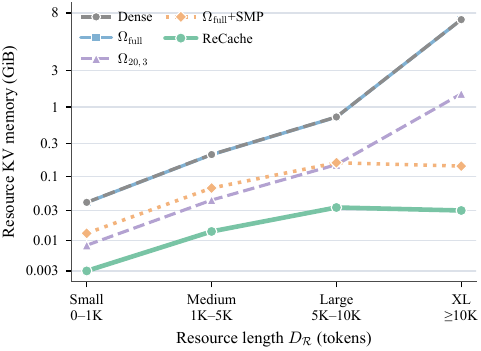}
        \caption{\small Mem.}
        \label{fig:res-eff-kv}
    \end{subfigure}
    \caption{\small Efficiency on varying resource lengths $D_{\mathcal{R}}$. Requests are grouped by average resource length into Small ($0 \leq D_{\mathcal{R}} < 1\mathrm{K}$), Medium ($1\mathrm{K} \leq D_{\mathcal{R}} < 5\mathrm{K}$), Large ($5\mathrm{K} \leq D_{\mathcal{R}} < 10\mathrm{K}$), and XL ($D_{\mathcal{R}} \geq 10\mathrm{K}$). SMP denotes semantic pruning.}
    \label{fig:res-eff}
\end{figure*}

We further evaluate efficiency across four ranges of total resource length $D_{\mathcal{R}}$. As shown in Figure~\ref{fig:res-eff}, Dense TTFT increases from tens of milliseconds to over 5,000 ms as resource contexts expand, whereas $\Omega_{\mathrm{full}}$ maintains TTFT below 50 ms through resource-level cache reuse. ReCache further reduces this prefill overhead, exhibiting an almost flat profile slightly above 5 ms across all ranges. 

Structural and semantic pruning further improve decoding efficiency. Dense and $\Omega_{\mathrm{full}}$ reach approximately 11 ms TPOT and 5 ms attention latency, whereas ReCache maintains TPOT slightly above 6 ms and attention latency below 0.2 ms with consistently stable decoding latency across resource scales. At XL, semantic pruning ($\Omega_{\mathrm{full}}$+SMP) reduces attention latency below 0.5 ms compared with over 2 ms under structural pruning ($\Omega_{20,3}$) alone, as longer resource contexts amplify the benefit of reducing tokens from semantic fields. Combining both pruning strategies therefore achieves the lowest and most stable latency.

ReCache also caps allocated Mem. at 0.03 GiB, whereas Dense and $\Omega_{\mathrm{full}}$ approach 8 GiB as resource contexts grow. These bounded prefill, decoding, and memory costs demonstrate the suitability of ReCache for agentic workloads with increasingly large resource contexts.

\section{Conclusion}
We presented ReCache, a framework for reusing and compressing tool and skill schemas in agentic language models. Resource-wise attention encodes each complete schema as a resource-composition-invariant KV block, eliminating repeated encoding across retrieval combinations. After construction, field-aware semantic pruning retains invocation-critical states, while contribution-guided structural pruning exposes them only to influential layer--KV-head-group routes. This separation preserves information during encoding while reducing the states stored and processed during inference.

Across seven tool- and skill-use datasets, independently reusable blocks match dense invocation quality and substantially reduced prefill latency. Combining structural and semantic pruning reduces allocated KV-tensor memory by 92.43\% and accelerates attention by 1.423$\times$, while retaining 97.5\% of dense Inv-F1 in-distribution and 91.8\% on unseen resources. The structural analysis further showed that effective routing budgets depend on model-scale redundancy, but can be selected with the same contribution-based criterion. ReCache therefore offers a practical way to scale dynamic resource contexts without repeatedly encoding or densely exposing every schema.

\section*{Limitations}
Our evaluation focuses on two Qwen3 backbones, with end-to-end experiments primarily conducted on Qwen3-4B, as full-parameter fine-tuning of larger models was beyond the computational budget of our current infrastructure. While the optimal structural budget varies across model scales, extending the analysis to larger models and other architectures remains an important direction. Since ReCache modifies attention masks and positional layouts, the current study focuses on trainable settings, while adapting it to frozen models is left for future work. In addition, ReCache is designed for environments with relatively stable system instructions and resource schemas, where resource representations can be reused across different user queries and retrieval combinations. Extending reuse to highly dynamic resource environments, where retrieval order or cross-resource dependencies carry essential information, provides another avenue for future exploration.


\bibliography{custom}

\clearpage
\appendix


\section{Dataset Construction Details}
\label{sec:app-dataset}

This appendix describes how the heterogeneous source pools are converted into the training and evaluation data used by ReCache. The objective is to obtain broad resource-use coverage without allowing invalid traces or repeated resource configurations to confound the evaluation of resource compression. Table~\ref{tab:dataset-details} reports the source-level statistics before and after construction.

\paragraph{Problems in direct aggregation.}
We combine ToolACE~\citep{liu2024toolace}, APIGEN~\citep{liu2024apigen}, ToolMind~\citep{yang2025toolmind}, Toucan~\citep{toucan2025}, ToolRet~\citep{toolret2025}, WildToolBench~\citep{yu2026benchmarking}, and SkillRouter~\citep{zheng2026skillrouter} because they cover complementary schema conventions, candidate-set sizes, interaction lengths, and resource types. Directly concatenating them, however, exposes two systematic problems. First, their declarations and invocations use heterogeneous formats, and generated trajectories may contain unparsable arguments, no actual invocation, or hallucinated invocations. Second, their resource distributions are highly repetitive. Across the five sources with frequency statistics, 29,232 of 37,824 source-level distinct resource names recur in multiple records (77.3\%), while 9,901 of 52,621 candidate records repeat an exact previously observed candidate subset (18.8\%). The imbalance is concentrated rather than uniform: recurring-name rates reach 99.96\% in APIGEN, 100\% in Toucan, and 99.95\% in ToolMind, while 52.8\% of Toucan records duplicate an existing candidate subset. Consequently, random downsampling would preserve shortcuts based on familiar resource identities and candidate configurations instead of encouraging resource-conditioned invocation.

\paragraph{Two-stage diversity-first pipeline.}
Before sampling, we normalize resource names by trimming whitespace and case, repair lightweight JSON-format errors, parse every declared resource and assistant invocation, and discard traces with no usable invocations, unresolved invocations, undeclared resource names, or sequences beyond the context budget. Repeated invocations to the same resource inside one trajectory are counted once, preventing retries from inflating frequency estimates; when source-provided quality labels are available, they are applied before sampling, including the realism, verifiability, stability, completeness, and conciseness filters for Toucan. The diversity sampler then proceeds in two stages. For each valid example $x$, we define the candidate-configuration signature
\begin{equation*}
\sigma(x)=\operatorname{sort}\{n(R):R\in\mathcal{R}(x)\},
\end{equation*}
where $\mathcal{R}(x)$ is its declared resource set and $n(R)$ is the normalized resource name.

In Stage~I, an example is accepted only if its signature has not appeared and none of its called resources has been used by a previously selected example, jointly prioritizing candidate-subset and invocation-level diversity. Since strict uniqueness cannot fill every source quota, Stage~II admits additional valid examples in the same quality-ranked order until each predefined quota is reached. This second stage preserves source and scenario coverage while retaining the diversity gains established by the first pass. A final global validity pass reduces the 52,964 sampled candidates to 49,424 training examples, removing 3,540 records (6.7\%).

\paragraph{Evaluation split construction.}
After the two-stage selection, we construct in-distribution (IND) and out-of-distribution (OOD) candidate pools at the resource level. For each resource $R_i$, $\operatorname{frq}(R_i)$ is the number of valid examples in which it is declared; low-frequency resources are reserved for OOD evaluation and removed from training. An example is eligible for the OOD pool only when every declared resource belongs to this reserved set, which prevents partial overlap with training resources. IND candidates use resources observed during training but contain different examples, queries, and invocation patterns. Exact configuration signatures are deduplicated independently within both pools before sampling, preventing repeated candidate subsets from making either test condition artificially easy. We finally sample 1,000 IND and 1,000 OOD examples, yielding matched evaluation sizes while isolating generalization to unseen resources.

\begin{table*}[t]
\centering
\small
\setlength{\tabcolsep}{4.0pt}
\begin{tabular}{llrrrrrr}
\toprule
\textbf{Group} & \textbf{Source} & \shortstack{\textbf{Train}\\\textbf{pool}} & \shortstack{\textbf{Distinct}\\\textbf{names}} & \shortstack{\textbf{Subset}\\\textbf{dup.}} & \shortstack{\textbf{Recurring}\\\textbf{names}} & \shortstack{\textbf{IND}\\\textbf{cand.}} & \shortstack{\textbf{OOD}\\\textbf{cand.}} \\
\midrule
\multirow{9}{*}{Tool}
& ToolACE & 8,400 & 14,566 & 178 & 7,144 & 200 & 195 \\
& APIGEN & 10,000 & 2,806 & 734 & 2,805 & 500 & 146 \\
& Toucan (all generators) & 13,071 & 4,355 & 6,901 & 4,355 & 600 & 493 \\
& \quad - Kimi-K2 & 3,333 & 1,103 & 1,772 & 1,103 & 200 & 145 \\
& \quad - GPT-OSS-120B & 4,184 & 1,412 & 2,162 & 1,412 & 200 & 130 \\
& \quad - Qwen3-32B & 5,554 & 1,840 & 2,967 & 1,840 & 200 & 218 \\
& ToolRet & 1,150 & 1,593 & 235 & 431 & 100 & 150 \\
& ToolMind & 20,000 & 14,504 & 1,853 & 14,497 & 500 & 190 \\
& WildToolBench & 256 & -- & -- & -- & -- & -- \\
\midrule
Skill & SkillRouter & 87 & -- & -- & -- & -- & -- \\
\midrule
\multicolumn{2}{l}{Sources with frequency statistics} & 52,621 & 37,824 & 9,901 & 29,232 & 1,900 & 1,174 \\
\multicolumn{2}{l}{Candidate pool} & 52,964 & -- & -- & -- & 1,900 & 1,174 \\
\multicolumn{2}{l}{Final dataset} & 49,424 & -- & -- & -- & 1,000 & 1,000 \\
\bottomrule
\end{tabular}
\caption{Dataset composition and split statistics. Source rows report eligible examples before final selection; the Toucan aggregate is not double-counted in the summary row. \textbf{Distinct names} counts source-level unique declared tools or skills, \textbf{Subset dup.} counts records beyond the first occurrence of an identical candidate configuration, and \textbf{Recurring names} counts resources appearing in multiple records. The summary row corresponds to 18.8\% duplicated candidate records and 77.3\% recurring resource names. IND cand. and OOD cand. denote eligible in-distribution and out-of-distribution candidates, and dashes mark unavailable statistics.}
\label{tab:dataset-details}
\end{table*}

\section{Structural Capacity Analysis}
\label{sec:app-budget}
\subsection{Preliminary Budget Investigation}
We investigate how computation is allocated across transformer layers and KV head groups under different model scales and training regimes when identifying the optimal structural budget $\Omega^\star$. The analysis focuses on two factors that influence the achievable sparsity level: the amount of optimization supervision and the representational capacity of the backbone model.

\paragraph{Observation 1: Head-group pruning is more sensitive to limited supervision.}

Under $\mathcal{D}_{\mathrm{subset}}$, $Q_s$ is substantially more sensitive to head-group sparsification than to layer sparsification. Retaining six of eight KV head groups across all layers, denoted by $\Omega_{L,6}$, reduces Inv-F1 by 18.7\% compared to $\Omega_{\mathrm{full}}$ (Table~\ref{tab:str-metrics-ob1-17b}). In contrast, retaining 20 of 28 layers across all head groups, denoted by $\Omega_{20,G}$, changes Inv-F1 by less than 0.2\%. This asymmetry indicates that head-group sparsification requires stronger supervision to identify effective resource-conditioned computation patterns.

The auxiliary-loss experiment provides further evidence. Incorporating an invocation-oriented objective for $Q_s$ with $\Omega_{L,6}$ under $\mathcal{D}_{\mathrm{subset}}$ increases Inv-F1 to 69.9\%, indicating that direct invocation supervision partially mitigates the degradation caused by head-group sparsification. Similarly, training on the full dataset $\mathcal{D}$ narrows the performance gap between sparse configurations and $\Omega_{\mathrm{full}}$. Together, these results show that the viable structural budget depends jointly on retained route capacity and the available optimization signal.

\begin{table*}[t]
\centering
\small
\setlength{\tabcolsep}{4pt}
\renewcommand{\arraystretch}{0.95}
\begin{minipage}[t]{0.49\textwidth}
\centering
$\boldsymbol{{\mathcal{D_{\operatorname{subset}}}}}$\\[2pt]
\begin{tabular}{lccccc}
\toprule
\textbf{Config} & \textbf{Inv-F1} & \multicolumn{3}{c}{\textbf{Resource ID}} & \textbf{Halluc.} \\
\cmidrule(lr){3-5}
& & \textbf{ID-P} & \textbf{ID-R} & \textbf{ID-F1} & \\
\midrule
$\Omega_{\mathrm{full}}$ & 72.1 & 92.0 & 90.8 & 90.9 & 0.4 \\
$\Omega_{L,5}$ & 49.6 & 74.5 & 73.5 & 73.4 & 4.8 \\
$\Omega_{L,6}$ & 53.4 & 79.6 & 77.9 & 78.3 & 3.2 \\
$\Omega_{L,7}$ & 57.3 & 83.4 & 81.8 & 82.1 & 2.9 \\
$\Omega_{15,G}$ & 69.3 & 89.2 & 88.3 & 88.3 & 1.1 \\
$\Omega_{20,G}$ & 70.5 & 90.6 & 89.5 & 89.6 & 1.1 \\
\bottomrule
\end{tabular}
\end{minipage}
\hfill
\begin{minipage}[t]{0.49\textwidth}
\centering
$\boldsymbol{\mathcal{D}}$\\[2pt]
\begin{tabular}{lccccc}
\toprule
\textbf{Config} & \textbf{Inv-F1} & \multicolumn{3}{c}{\textbf{Resource ID}} & \textbf{Halluc.} \\
\cmidrule(lr){3-5}
& & \textbf{ID-P} & \textbf{ID-R} & \textbf{ID-F1} & \\
\midrule
$\Omega_{\mathrm{full}}$ & 80.7 & 94.9 & 94.3 & 94.4 & 0.1 \\
$\Omega_{L,5}$ & 79.5 & 94.4 & 93.6 & 93.8 & 0.6 \\
$\Omega_{L,6}$ & 79.8 & 94.4 & 93.9 & 93.9 & 0.3 \\
$\Omega_{L,7}$ & 80.6 & 94.8 & 94.2 & 94.3 & 0.2 \\
$\Omega_{15,G}$ & 80.0 & 94.2 & 93.7 & 93.8 & 0.3 \\
$\Omega_{20,G}$ & 80.6 & 94.7 & 94.1 & 94.2 & 0.4 \\
\bottomrule
\end{tabular}
\end{minipage}
\caption{\small Structural budget ablation for $Q_s$ across training data regimes with comparable budgets.}
\label{tab:str-metrics-ob1-17b}
\end{table*}

\paragraph{Observation 2: Model scale affects the required structural budget.}

Despite full-data training, models with different capacities retain distinct structural requirements (Table~\ref{tab:str-metrics-ob2}). $Q_l$ maintains comparable invocation performance with fewer active head groups, while $Q_s$ requires broader head-group participation. This implies that larger models can sustain resource-conditioned computation with more aggressive structural sparsification.

\begin{table}[ht]
\centering
\small
\setlength{\tabcolsep}{2pt}
\renewcommand{\arraystretch}{0.95}
\begin{tabular}{@{}lccccc@{}}
\toprule
\textbf{Config} & \textbf{Inv-F1} & \multicolumn{3}{c}{\textbf{Resource ID}} & \textbf{Halluc.} \\
\cmidrule(lr){3-5}
& & \textbf{ID-P} & \textbf{ID-R} & \textbf{ID-F1} & \\
\midrule
\multicolumn{6}{@{}l@{}}{$\boldsymbol{Q_s}$} \\
\addlinespace[2pt]
$\Omega_{\mathrm{full}}$ & 80.7 & 94.9 & 94.3 & 94.4 & 0.1 \\
$\Omega_{20,5}$ & 78.8 & 94.1 & 93.5 & 93.6 & 0.5 \\
$\Omega_{20,7}$ & 80.0 & 94.7 & 94.0 & 94.1 & 0.3 \\
\midrule
\multicolumn{6}{@{}l@{}}{$\boldsymbol{Q_l}$} \\
\addlinespace[2pt]
$\Omega_{\mathrm{full}}$ & 82.3 & 96.4 & 95.8 & 96.0 & 0.2 \\
$\Omega_{20,2}$ & 81.9 & 95.3 & 94.7 & 94.7 & 0.3 \\
$\Omega_{20,3}$ & 82.1 & 96.1 & 95.5 & 95.6 & 0.2 \\
\bottomrule
\end{tabular}
\caption{\small Structural budget ablation trained on $\mathcal{D}$.}
\label{tab:str-metrics-ob2}
\end{table}

Overall, these observations suggest that the attainable structural sparsity is determined by the effective capacity of retained resource-visible routes, which depends on both optimization supervision and model representation capability. The larger Qwen3-4B provides greater capacity within each retained route, attributable to its larger hidden and FFN dimensions and its four query heads per KV group, compared with two in Qwen3-1.7B.

\subsection{Contribution Distribution}
\label{sec:app-weights}

Table~\ref{tab:app-wi} reports detailed weights of individual units, the cumulative contribution coverage of layers, and head groups, respectively.

\section{Semantic Pruning Ablation}
\label{sec:app-sm}
Table~\ref{tab:sm-metrics-4b} presents the semantic pruning ablation on $\mathcal{D}_{\mathrm{subset}}$, where all experiments are conducted under $\Omega_{20,3}$. The objective is to identify the semantic fields that should be retained before scaling the selected configuration to full-data training.

The results reveal a clear hierarchy among the evaluated semantic fields. Adding only the resource name already yields substantial improvements, reducing the hallucination rate by 97.63\% relative to the suffix-only setting while increasing all resource-ID metrics above 88\%. Incorporating argument names further improves Inv-F1 by 22.5\%, demonstrating that parameter identifiers provide essential grounding for argument generation.

\begin{table}[!ht]
\centering
\small
\setlength{\tabcolsep}{1pt}
\renewcommand{\arraystretch}{0.95}
\begin{tabular}{lccccc}
\toprule
\textbf{Config} & \textbf{Inv-F1} & \multicolumn{3}{c}{\textbf{Resource ID}} & \textbf{Halluc.} \\
\cmidrule(lr){3-5}
& & \textbf{ID-P} & \textbf{ID-R} & \textbf{ID-F1} & \\
\midrule
$\Omega_{20,3}$  & 76.1 & 93.1 & 92.8 & 92.5 & 0.9 \\

\multicolumn{6}{l}{\textit{Schema representation ablation}}\\
Suffix-only Schema & 14.8 & 22.8 & 23.0 & 22.9 & 76.0 \\
\quad + Resource Name & 45.2 & 89.9 & 88.8 & 88.9 & 1.8 \\
\quad\quad + Arg Name & 67.7 & 92.1 & 91.1 & 91.2 & 1.3 \\
\quad\quad\quad + Resource Desc. & 67.9 & 92.2 & 91.4 & 91.4 & 1.1 \\
\quad\quad\quad + Arg Desc. & 72.8 & 92.8 & 91.9 & 92.0 & 1.0 \\
\quad\quad\quad + Both Desc. & 73.4 & 93.0 & 92.2 & 92.2 & 1.2 \\
\bottomrule
\end{tabular}
\caption{\small Semantic pruning ablation trained on $\mathcal{D}_{\mathrm{subset}}$. Arg stands for argument, Desc. stands for description. $\Omega_{20,3}$ denotes the routing configuration with $K_L=20,K_G=3$.}
\label{tab:sm-metrics-4b}
\end{table}

Within the considered schema components, argument descriptions provide additional benefits by supplying semantic constraints for parameter prediction. In contrast, resource descriptions offer limited contribution. Adding only resource descriptions on top of resource and argument names yields negligible improvement, while additionally including resource descriptions alongside argument descriptions changes all evaluation metrics by less than 0.7\% compared with preserving argument descriptions alone. These results suggest that resource names, argument names, and argument descriptions constitute the core semantic components for resource invocation, while resource descriptions provide limited additional information once the executable interface has been preserved.

\begin{table*}[t]
    \centering
    \footnotesize
    \setlength{\tabcolsep}{3pt}
    \renewcommand{\arraystretch}{0.92}
    \begin{tabular}{@{}r l r r @{\hspace{1.5em}} r l r r@{}}
    \toprule
    \multicolumn{8}{c}{\textbf{Layers}} \\
    \addlinespace[1pt]
    & \multicolumn{3}{c}{$\boldsymbol{Q_l}$\textbf{(Qwen3-4B)}} &
    & \multicolumn{3}{c}{$\boldsymbol{Q_s}$\textbf{(Qwen3-1.7B)}} \\
    \cmidrule(lr){2-4}\cmidrule(lr){6-8}
    \textbf{rank} & \textbf{unit} & $\boldsymbol{w_i}$ (\%) & \textbf{CC} (\%) &
    \textbf{rank} & \textbf{unit} & $\boldsymbol{w_i}$ (\%) & \textbf{CC} (\%) \\
    \midrule
    \textbf{1}  & \textbf{$L_{34}$} & \textbf{15.56} & \textbf{15.6} &
    \textbf{1}  & \textbf{$L_{22}$} & \textbf{20.09} & \textbf{20.1} \\
    \textbf{2}  & \textbf{$L_{35}$} & \textbf{10.53} & \textbf{26.1} &
    \textbf{2}  & \textbf{$L_{25}$} & \textbf{12.90} & \textbf{33.0} \\
    \textbf{3}  & \textbf{$L_{23}$} & \textbf{9.37}  & \textbf{35.5} &
    \textbf{3}  & \textbf{$L_{20}$} & \textbf{12.14} & \textbf{45.1} \\
    \textbf{4}  & \textbf{$L_{25}$} & \textbf{8.31}  & \textbf{43.8} &
    \textbf{4}  & \textbf{$L_{26}$} & \textbf{10.90} & \textbf{56.0} \\
    \textbf{5}  & \textbf{$L_{32}$} & \textbf{8.01}  & \textbf{51.8} &
    \textbf{5}  & \textbf{$L_{27}$} & \textbf{10.39} & \textbf{66.4} \\
    \textbf{6}  & \textbf{$L_{30}$} & \textbf{6.95}  & \textbf{58.7} &
    \textbf{6}  & \textbf{$L_{17}$} & \textbf{6.82}  & \textbf{73.2} \\
    \textbf{7}  & \textbf{$L_{33}$} & \textbf{6.89}  & \textbf{65.6} &
    \textbf{7}  & \textbf{$L_{21}$} & \textbf{6.70}  & \textbf{79.9} \\
    \textbf{8}  & \textbf{$L_{27}$} & \textbf{5.17}  & \textbf{70.8} &
    \textbf{8}  & \textbf{$L_{19}$} & \textbf{4.65}  & \textbf{84.6} \\
    \textbf{9}  & \textbf{$L_{21}$} & \textbf{4.87}  & \textbf{75.7} &
    \textbf{9}  & \textbf{$L_{24}$} & \textbf{4.00}  & \textbf{88.6} \\
    \textbf{10} & \textbf{$L_{28}$} & \textbf{4.64}  & \textbf{80.3} &
    \textbf{10} & \textbf{$L_{23}$} & \textbf{2.13}  & \textbf{90.7} \\
    \textbf{11} & \textbf{$L_{24}$} & \textbf{4.02}  & \textbf{84.3} &
    \textbf{11} & \textbf{$L_{18}$} & \textbf{1.89}  & \textbf{92.6} \\
    \textbf{12} & \textbf{$L_{22}$} & \textbf{3.74}  & \textbf{88.1} &
    \textbf{12} & \textbf{$L_{15}$} & \textbf{1.73}  & \textbf{94.3} \\
    \textbf{13} & \textbf{$L_{17}$} & \textbf{2.30}  & \textbf{90.4} &
    \textbf{13} & \textbf{$L_{13}$} & \textbf{1.39}  & \textbf{95.7} \\
    \textbf{14} & \textbf{$L_{31}$} & \textbf{1.96}  & \textbf{92.3} &
    \textbf{14} & \textbf{$L_{10}$} & \textbf{1.13}  & \textbf{96.9} \\
    \textbf{15} & \textbf{$L_{18}$} & \textbf{1.29}  & \textbf{93.6} &
    \textbf{15} & \textbf{$L_{9}$}  & \textbf{1.03}  & \textbf{97.9} \\
    \textbf{16} & \textbf{$L_{14}$} & \textbf{1.26}  & \textbf{94.9} &
    \textbf{16} & \textbf{$L_{4}$}  & \textbf{0.88}  & \textbf{98.8} \\
    \textbf{17} & \textbf{$L_{20}$} & \textbf{0.78}  & \textbf{95.7} &
    \textbf{17} & \textbf{$L_{16}$} & \textbf{0.56}  & \textbf{99.3} \\
    \textbf{18} & \textbf{$L_{36}$} & \textbf{0.69}  & \textbf{96.3} &
    \textbf{18} & \textbf{$L_{12}$} & \textbf{0.18}  & \textbf{99.5} \\
    \textbf{19} & \textbf{$L_{19}$} & \textbf{0.68}  & \textbf{97.0} &
    \textbf{19} & \textbf{$L_{11}$} & \textbf{0.15}  & \textbf{99.7} \\
    \textbf{20} & \textbf{$L_{16}$} & \textbf{0.68}  & \textbf{97.7} &
    \textbf{20} & \textbf{$L_{8}$}  & \textbf{0.14}  & \textbf{99.8} \\
    \midrule
    21 & $L_{10}$ & 0.60 & 98.3 & 21 & $L_{28}$ & 0.11 & 99.9 \\
    22 & $L_{5}$  & 0.33 & 98.6 & 22 & $L_{5}$  & 0.06 & 100.0 \\
    23 & $L_{11}$ & 0.29 & 98.9 & 23 & $L_{2}$  & 0.02 & 100.0 \\
    24 & $L_{9}$  & 0.27 & 99.2 & 24 & $L_{1}$  & 0.00 & 100.0 \\
    25 & $L_{12}$ & 0.26 & 99.5 & 25 & $L_{3}$  & 0.00 & 100.0 \\
    26 & $L_{15}$ & 0.26 & 99.7 & 26 & $L_{6}$  & 0.00 & 100.0 \\
    27 & $L_{13}$ & 0.13 & 99.8 & 27 & $L_{7}$  & 0.00 & 100.0 \\
    28 & $L_{8}$  & 0.07 & 99.9 & 28 & $L_{14}$ & 0.00 & 100.0 \\
    29 & $L_{3}$  & 0.07 & 100.0 & -- & -- & -- & -- \\
    30 & $L_{1}$  & 0.00 & 100.0 & -- & -- & -- & -- \\
    31 & $L_{2}$  & 0.00 & 100.0 & -- & -- & -- & -- \\
    32 & $L_{4}$  & 0.00 & 100.0 & -- & -- & -- & -- \\
    33 & $L_{6}$  & 0.00 & 100.0 & -- & -- & -- & -- \\
    34 & $L_{7}$  & 0.00 & 100.0 & -- & -- & -- & -- \\
    35 & $L_{26}$ & 0.00 & 100.0 & -- & -- & -- & -- \\
    36 & $L_{29}$ & 0.00 & 100.0 & -- & -- & -- & -- \\
    \midrule
    \multicolumn{8}{c}{\textbf{KV head groups}} \\
    \addlinespace[1pt]
    & \multicolumn{3}{c}{$\boldsymbol{Q_l}$\textbf{(Qwen3-4B)}} &
    & \multicolumn{3}{c}{$\boldsymbol{Q_s}$\textbf{(Qwen3-1.7B)}} \\
    \cmidrule(lr){2-4}\cmidrule(lr){6-8}
    \textbf{rank} & \textbf{unit} & $\boldsymbol{w_i}$ (\%) & \textbf{CC} (\%) &
    \textbf{rank} & \textbf{unit} & $\boldsymbol{w_i}$ (\%) & \textbf{CC} (\%) \\
    \midrule
    \textbf{1} & \textbf{$G_{4}$} & \textbf{16.54} & \textbf{16.5} &
    \textbf{1} & \textbf{$G_{5}$} & \textbf{22.84} & \textbf{22.8} \\
    \textbf{2} & \textbf{$G_{6}$} & \textbf{16.17} & \textbf{32.7} &
    \textbf{2} & \textbf{$G_{3}$} & \textbf{19.01} & \textbf{41.9} \\
    \textbf{3} & \textbf{$G_{3}$} & \textbf{14.64} & \textbf{47.3} &
    \textbf{3} & \textbf{$G_{8}$} & \textbf{17.27} & \textbf{59.1} \\
    \cmidrule(r){1-4}
    4 & $G_{2}$ & 12.37 & 59.7 &
    \textbf{4} & \textbf{$G_{6}$} & \textbf{16.77} & \textbf{75.9} \\
    5 & $G_{8}$ & 11.62 & 71.3 &
    \textbf{5} & \textbf{$G_{1}$} & \textbf{10.91} & \textbf{86.8} \\
    6 & $G_{1}$ & 9.68  & 81.0 &
    \textbf{6} & \textbf{$G_{7}$} & \textbf{8.83}  & \textbf{95.6} \\
    7 & $G_{5}$ & 9.61  & 90.6 &
    \textbf{7} & \textbf{$G_{4}$} & \textbf{3.64}  & \textbf{99.3} \\
    \cmidrule(l){5-8}
    8 & $G_{7}$ & 9.37  & 100.0 &
    8 & $G_{2}$ & 0.74  & 100.0 \\
    \bottomrule
    \end{tabular}
    \caption{\small Individual contribution weights and cumulative contribution
    coverage of structural units. Within each panel, units are ranked independently
    by decreasing $w_i$, and CC at rank $k$ is the cumulative positive contribution
    weight through rank $k$. Bold entries fall within the empirical budgets
    $K_L{=}20,K_G{=}3$ for $Q_l$ and $K_L{=}20,K_G{=}7$ for $Q_s$.}
    \label{tab:app-wi}
\end{table*}

\end{document}